\documentclass{article}
\usepackage{ijcai23}

\usepackage{times}
\usepackage{soul}
\usepackage{url}
\usepackage[hidelinks]{hyperref}
\usepackage[utf8]{inputenc}
\usepackage[small]{caption}
\usepackage{graphicx}
\usepackage{amsmath}
\usepackage{amsthm}
\usepackage{booktabs}
\usepackage{algorithm}
\usepackage{algorithmic}
\usepackage{amssymb}
\usepackage[switch]{lineno}
\usepackage{threeparttable}
\usepackage{multirow}
\usepackage[table]{xcolor}
\usepackage[a-1b]{pdfx}
\title{A Low Latency Adaptive Coding Spike Framework for Deep Reinforcement
Learning}

\author{
Lang Qin$^1$
\and
Rui Yan$^2$\And
Huajin Tang$^{1,3}$\thanks{Corresponding author}\\
\affiliations
$^1$College of Computer Science and Technology, Zhejiang University, Hangzhou, China\\
$^2$College of Computer Science, Zhejiang University of Technology, Hangzhou, China\\
$^3$Zhejiang Lab, Hangzhou, China\\
\emails
qinl@zju.edu.cn,
ryan@zjut.edu.cn,
htang@zju.edu.cn
}

\begin{document}
\maketitle

\begin{abstract}
    In recent years, spiking neural networks (SNNs) have been used in reinforcement learning (RL) due to their low power consumption and event-driven features. However, spiking reinforcement learning (SRL), which suffers from fixed coding methods, still faces the problems of high latency and poor versatility. In this paper, we use learnable matrix multiplication to encode and decode spikes, improving the flexibility of the coders and thus reducing latency. Meanwhile, we train the SNNs using the direct training method and use two different structures for online and offline RL algorithms, which gives our model a wider range of applications. Extensive experiments have revealed that our method achieves optimal performance with ultra-low latency (as low as 0.8\% of other SRL methods) and excellent energy efficiency (up to 5X the DNNs) in different algorithms and different environments.
\end{abstract}

\section{Introduction}
Deep learning has had a significant impact on many areas of machine learning, including reinforcement learning (RL). The combination of deep learning algorithms and RL has led to the development of a new field called deep reinforcement learning (DRL) \cite{arulkumaran2017brief}, which has achieved impressive results in a variety of RL tasks, including some that have reached or even surpassed human-level performance \cite{mnih2015human,DBLP:conf/iclr/HuZCL21,DBLP:journals/nature/SilverHMGSDSAPL16}. However, DRL using deep neural networks (DNNs) requires a significant amount of resources, which may not always be available. For example, in the case of mobile robots, the processing system needs to be power-efficient and have low latency (inference time) \cite{niroui2019deep,lahijanian2018resource}. As a result, researchers are actively exploring the use of alternative networks for DRL that are more energy-efficient and have lower latency.

Inspired by neurobiology, spiking neural networks (SNNs) use differential dynamics equations and spike information encoding methods to build computing node models in neural networks \cite{maass1997networks}. Compared with DNNs, SNNs are closer to biological neural networks and could greatly reduce energy consumption with neuromorphic hardware. Therefore, SNNs are well suited as a low-power alternative to DNNs in DRL. But there is still relatively little attempt at using SNNs in RL to achieve high-dimensional control with low latency (inference time steps) and energy consumption. 

The existing spiking reinforcement learning (SRL) methods are mainly divided into two categories \cite{DBLP:journals/natmi/NeftciA19}. First, guided by revealing reward mechanisms in the brain, several works \cite{yuan2019reinforcement,bing2018end,fremaux2013reinforcement} train spiking neurons with reward-based local learning rules. These local learning methods can only be applied for simple control with shallow SNNs. Second, oriented toward better performance, replacing DNNs in DRL with SNNs can also implement SRL. Since SNNs usually use the firing rate as the equivalent activation value \cite{DBLP:conf/nips/LiGZDHG21}, which is a discrete value between 0 and 1, it is hard to represent the value function of RL, which does not have a certain range in training. To avoid using discrete spikes to calculate the value function, this type of method is subdivided into three types. Tan \shortcite{tan2021strategy} and Patel \shortcite{patel2019improved} use the SNN conversion method to convert the Deep Q-Networks (DQN) \cite{mnih2015human} into SNNs. Zhang \shortcite{zhang2022multiscale} and Tang \shortcite{tang2020reinforcement} use the hybrid framework, which uses DNNs for value function estimation and SNNs for execution. Chen \shortcite{chen2022deep} and Liu \shortcite{liu2021human} use surrogate gradients to train SNNs directly and fixed spike coders to estimate the value function.

\begin{figure*}[t]
	\centerline{\includegraphics[width=1.0\linewidth]{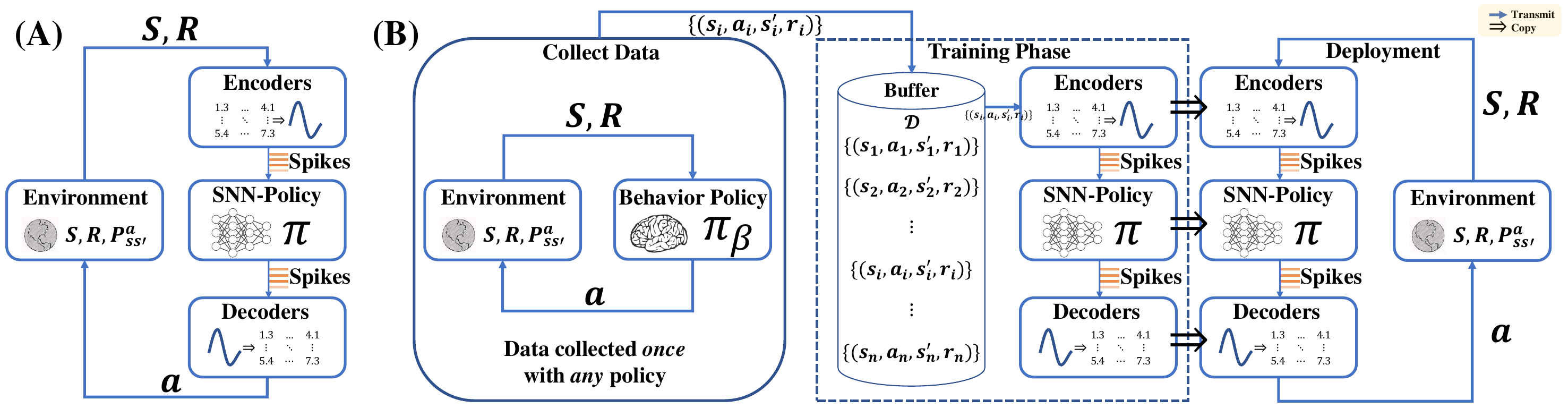}}
	\caption{Online and offline SRL frameworks. Environments generally contain elements such as states ($S$), rewards ($R$), and state transition probabilities ($P_{ss'}^a$). The state ($S$) is transmitted to the SNNs through the encoders. Action ($a$) and value functions are expressed by the decoders. The environment accepts the action ($a$) and gives the next state ($S'$) based on the transition probability ($P_{ss'}^a$). (A) In online SRL, the SNN-based policy $\pi$ interacts directly with the environment. (B) In offline SRL, the SNN-based policy interacts with a dataset $\mathcal{D}$ which collected by the behavior policy $\pi_{\beta}$. The behavior policy are usually developed by experienced humans or well-trained agent.}
	\label{Fig::architecture}
\end{figure*}

However, these methods all have their shortcomings. Conversion methods have low energy efficiency and high latency, while the hybrid framework loses the potential biological plausibility of SNNs and limits the application of different types of RL algorithms. Although the directly trained SNNs have high energy efficiency and low latency, they are hard to estimate for continuous-valued functions using fixed coders. They have to find a trade-off between latency and accuracy \cite{Li_2022_CVPR}. In addition, these SRL methods have a narrow range of applications because they are all limited to a specific online RL algorithm (DQN, TD3 \cite{DBLP:conf/icml/FujimotoHM18}, and DDPG \cite{lillicrap2015continuous}).

To overcome these shortcomings, we propose an adaptive coding method to compress or extend information in the temporal dimension through learnable matrix multiplication. This learnable code offers increased flexibility over fixed coding, allowing for better results with lower latency. Unlike sequence-to-sequence coding in \cite{DBLP:conf/nips/ShresthaO18} and \cite{DBLP:conf/nips/HagenaarsPC21}, our proposed adaptive coder implements value-sequence coding. Moreover, our coders do not require a label signal and can automatically select the optimal form according to different tasks.

At the same time, in order to expand the scope of application, we design structures for online and offline SRL algorithms, respectively. Figure~\ref{Fig::architecture}A shows the flow of the online algorithm, where the state and reward of the environment are directly interacted with the spiking agent after passing through the encoder. SNN calculates the value function and outputs it in the form of spikes, which are converted into real values by decoders. Finally, the action selection is made according to the value function. Offline algorithms are more complex (Figure~\ref{Fig::architecture}B). It begins by collecting data through behavioral policies that interact with the environment. Then, the spiking agent interacts with the collected data and learns through the same process as online algorithms. We named the proposed SRL model adaptive coding spike framework (ACSF). The main contributions of this paper can be summarized as follows:

\begin{itemize}
    \item We propose an adaptive coding spike framework for SRL that uses adaptive coders to represent states and estimate value functions. This framework reduces the latency of existing methods effectively while maintaining optimal performance.
    \item To our best knowledge, the ACSF is the first complete, directly trained SRL framework. It supports both online and offline RL algorithms, expanding the application range of SRL. 
    \item For the RL scale, ACSF provides a low-power inference scheme for RL algorithms (up to 5X energy efficient). This provides potential assistance for mobile robot control using RL algorithms.
    \item From the experimental results, compared with conversion and other directly trained methods, ACSF has better versatility and lower latency ($T=4$). At the same time, it achieves the best performance in all conditions.
\end{itemize}

\section{Related Works}

\begin{figure*}[t]
	\centerline{\includegraphics[width=1.0\linewidth]{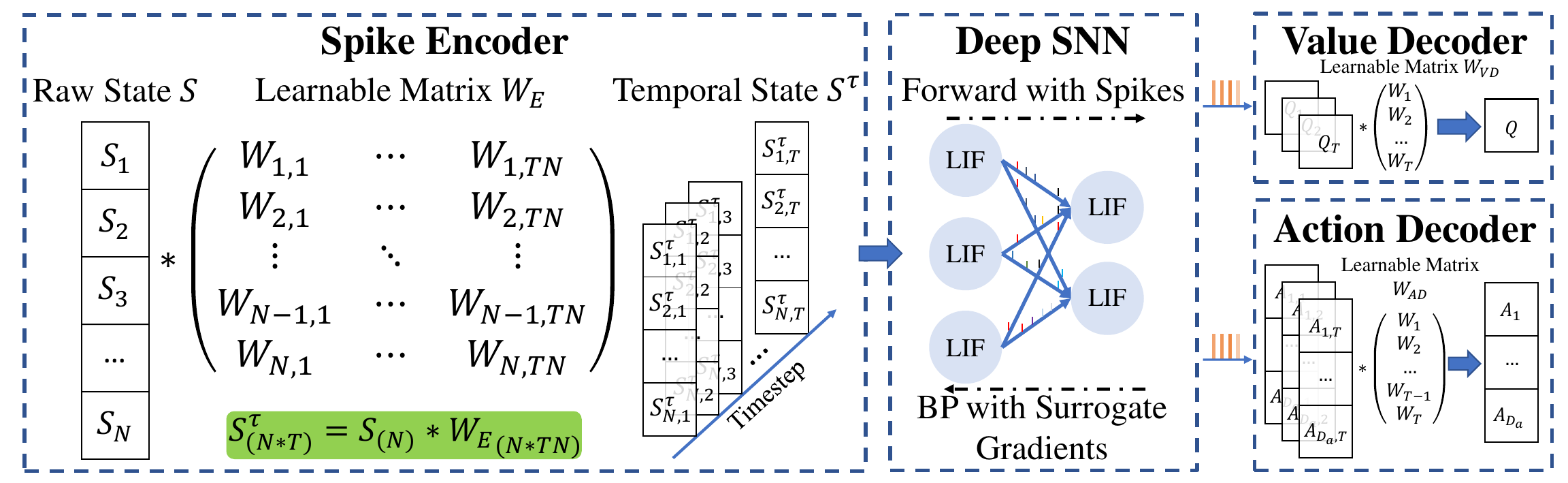}}
	\caption{The overall structure and workflow of the ACSF. The encoder transforms the raw state $S$ into the temporal state $S^\tau$, which is then fed into SNNs. The output spike trains generated by SNNs are decoded into values or actions by different decoders. Both the spike encoder and the decoder use learnable matrix multiplication to expand or compress inputs in the time dimension. Deep SNNs are trained directly using surrogate gradients.}
	\label{Fig::ACSFStructure}
\end{figure*}

\subsection{Reward-based Local Learning}
To explore the reward mechanism in the brain, applying reward-based training rules to the local learning of spiking neurons \cite{yuan2019reinforcement,bing2018end,fremaux2013reinforcement,legenstein2008learning} was the mainstream approach for early SRL. However, they have the common problem that they are limited to shallow SNNs and are hard to apply to complex tasks. E-prop \cite{bellec2020solution} successfully advances SRL to complex tasks with good performance. However, it needs to tune a lot of hyperparameters, making it difficult to apply to different RL tasks.

\subsection{Convert DNNs to SNNs for RL}
The conversion algorithms map the DNNs to the SNNs by matching the firing rate and activation values. It is a commonly used method for obtaining high-performance SNNs. Several works \cite{tan2021strategy,patel2019improved} obtain Deep Spiking Q-Networks (DSQN) using conversion methods. To ensure the accuracy of mapping, those methods often have high latency (about 500 time steps) and cannot outperform the source DQNs. 

\subsection{Hybrid Framework of SNNs and DNNs}
In RL, some algorithms use the actor-critic architecture \cite{sutton1999policy}, where value function estimation and action selection are implemented by critic and actor networks, respectively. Some works \cite{zhang2022multiscale,tang2020reinforcement} use deep critic networks and spiking actor networks to implement SRL. These methods can be applied to complex tasks with lower latency, but they only work for actor-critic structures, and they rely on DNNs, which makes them difficult to fully apply to neuromorphic chips.

\subsection{Directly Trained SNNs for RL}
Surrogate gradient methods introduce backpropagation (BP) into SNNs successfully and make it easy to train deep SNNs directly. Recently, a few papers \cite{chen2022deep,liu2021human} have used directly trained SNNs and fixed coders to construct DSQN. Although these methods have high energy efficiency, they do not form a complete framework and have a poor trade-off between accuracy and latency. In addition, their applications are also limited to discrete action-space environments.

\section{Methods}
In this section, we first introduce the DRL algorithms and the spiking neuron model. Then, we describe the adaptive coding method in detail. Finally, we derive the gradient of the direct training method. The overall architecture of ACSF is shown in Figure~\ref{Fig::ACSFStructure}.

\subsection{DRL Algorithms}
In this section, we will introduce different baseline algorithms (DQN, DDPG, BCQ and behavioral cloning) used in ACSF. Due to space limitations, we only show the final loss functions. Please refer to the supplementary materials for more details. Assume the agent's current environment sampling result is state $s$, action $a$, reward $r$ and next state $s'$. Mark the reward discount factor as $\gamma$.

\subsubsection{DQN}
DQN is a typical RL algorithm, and it only needs to learn the value function ($q$-values). Its loss function \cite{girshick2015fast} is:
\begin{small}
\begin{equation} \label{eq::DQN}
    \begin{split}
        &smooth_{L_1} (x) = \left \{
        \begin{array}{ll}
            0.5x^2 & if|x|<1 \\
            |x| - 0.5 & otherwise
        \end{array}
		\right. \\
        &\mathcal{L}_Q = smooth_{L_1} [r+\gamma \max_{a'}Q_{\theta_2}(s',a')-Q_{\theta_1}(s,a)]
    \end{split}
\end{equation}
\end{small}
where $Q_{\theta_1}$ and $Q_{\theta_2}$ is the current and target $q$-networks, respectively. 
\subsubsection{DDPG}
DDPG is based on AC architecture, including actor networks that select actions and critic networks that calculate value functions. The loss functions of the actor and critic networks are:
\begin{small}
\begin{equation} \label{eq::DDPG}
    \begin{split}
        \mathcal{L}_A &= -\frac{1}{N}\sum_{j=1}^{N}Q_{\theta_1}(s,A_{\phi_1}(s))\\
        \mathcal{L}_Q &= \frac{1}{N}\sum_{j=1}^{N}[r+\gamma Q_{\theta_2}(s',A_{\phi_2}(s'))-Q_{\theta_1}(s,a)]^2
    \end{split}
\end{equation}
\end{small}
where $Q_{\theta_1}$ and $Q_{\theta_2}$ are current and target critic networks; $A_{\phi_1}$ and $A_{\phi_2}$ are current and target actor networks. $N$ is the batch size.
\subsubsection{BCQ}
BCQ is an offline RL algorithm. In addition to training double critic networks $Q^1_{\theta_1}, Q^2_{\theta_1}$, target critic networks $Q^1_{\theta_2}, Q^2_{\theta_2}$, current perturbation network $\xi_{\delta_1}$ and target perturbation network $\xi_{\delta_2}$, we also need to train a variational auto-encoder (VAE) \cite{kingma2013auto} $G_{\omega}=\left \{{E_{\omega_1},D_{\omega_2}} \right\}$. 
Their loss functions can be written as follows \cite{fujimoto2019off}:
\begin{small}
\begin{equation} 
    \begin{split}
        &\mathcal{L}_G = \sum(a-\widetilde{a})^2 + D_{KL}(\mathcal{N}(\mu,\sigma)||\mathcal{N}(0,1))\\
        &\mathcal{L}_Q = \frac{1}{2} \sum[y-Q^1_{\theta_1}(s,a)]^2 + \frac{1}{2} \sum[y-Q^2_{\theta_1}(s,a)]^2\\
        &\mathcal{L}_\xi = -\sum Q^1_{\theta_1}(s,a+\xi_{\delta_1}(s,a,\Phi)), a \sim G_\omega(s)
    \end{split}
\label{eq::BCQ}
\end{equation}
\end{small}

\subsubsection{Behavioral Cloning}
Behavioral cloning (BC) is an offline imitation learning method used to fit the distribution of buffered data $\mathcal{D}$. We use a multilayer perceptron (MLP) $P_\kappa$ in this work. The input of the MLP model is raw states ($s$), and the output is action ($a$). We use the mean squared error (MSE) as the loss function:
\begin{small}
\begin{equation} 
    \begin{split}
        \mathcal{L}_P=\frac{1}{2} \sum_{i=1}^m [P_\kappa(s)-a]^2
    \end{split}
\label{eq::BC}
\end{equation}
\end{small}
where $m$ is the dimension of action.

\subsection{Iterative LIF Model}
\begin{figure}[t]
	\centerline{\includegraphics[width=1.0\linewidth]{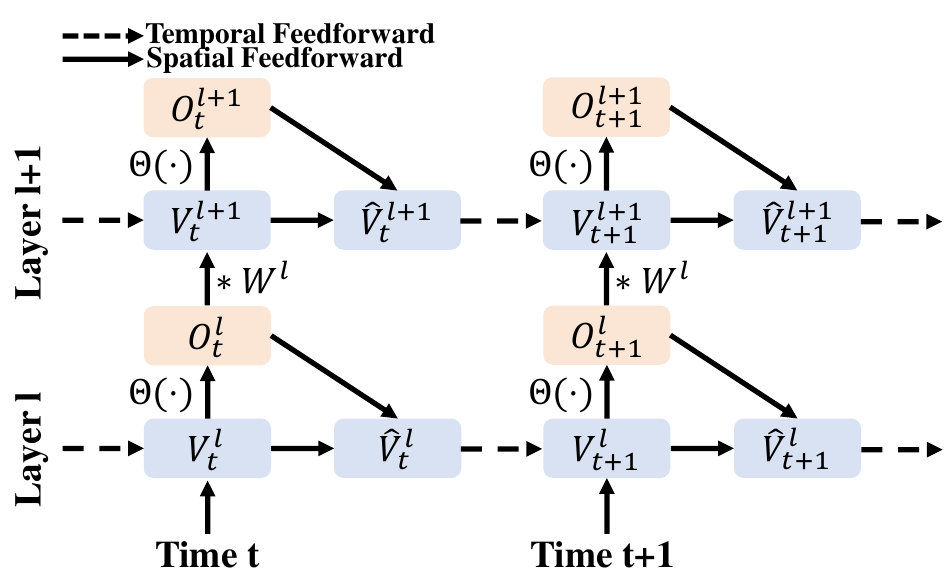}}
	\caption{Spatiotemporal dynamics of neurons. The solid arrows and the dotted arrows indicate the directions of spatial and temporal feedforward, respectively. The input signal changes the membrane potential ($V^l_t$) and fires the output spikes ($O^l_t$) through the processes of charging, discharging, and resetting.}
	\label{Fig::Neuron Dynamics}
\end{figure}

In this work, we employ the iterative leaky integrate-and-fire (LIF) neuron model to drive the proposed ACSF due to its biological plausibility and computational tractability. Its dynamics can be described by three processes: charge, discharge, and reset. The corresponding equations are:
\begin{small}
\begin{equation}
    \label{eq::LIF}
    \begin{aligned}
        V^l_t &= \beta \hat{V}^l_{t-1} + (1-\beta) W^lO^{l-1}_t & Charge\\
        O^l_t &= \Theta (V^l_t) , \ \Theta (x) = \left\{ \begin{array}{l} 
            1,x \ge \vartheta\\
            0,x < \vartheta
                        \end{array} \right. & Discharge\\
        \hat{V}^l_t &= V_rO^l_t + (1-O^l_t) V^l_t & Reset \\
    \end{aligned}
\end{equation}
\end{small}
where the superscript $l$ indicates the $l$-th layer, the subscript $t$ represents the $t$-th time step; $V$ and $\hat{V}$ denote the membrane potential before and after reset respectively; $\beta$ is the decay factor; $O$ is the output spike; $\Theta(\cdot)$ is the firing function; $V_r$ is the reset voltage; $\vartheta$ is the firing threshold. The schematic diagram of neuron dynamics is shown in Figure~\ref{Fig::Neuron Dynamics}.

\subsection{Adaptive Coders}
As mentioned earlier, existing directly trained SRL methods have limited application and poor trade-offs between performance and latency. The key to solving these problems is to design a common adaptive coder with learnable weights that allows the model to be adapted to different algorithms and environments. In addition, adaptive coding has better characterization capabilities than fixed coding and can achieve better performance with lower latency.

\subsubsection{Spike Encoder}
There are many fixed encoding methods in SRL, such as rate coding, population coding, and temporal coding. Although these methods have had some success, they are frequently rigid, with strict time step length requirements. To increase flexibility and reduce latency, we encode the raw state $S$ with a learnable matrix $W_E$ (Figure~\ref{Fig::ACSFStructure}). We use matrix multiplication to expand $S$ in the temporal dimension by $T$ times, yielding the temporal state $S^\tau$. The temporal state $S^\tau$ is then fed into the spiking neurons, which produce spike trains $O_t$. The encoding strategy at the $t$-th time step can be stated as follows:
\begin{small}
\begin{equation}\label{eq::Encoder}
	\begin{aligned}
        & S^\tau = S * W_E\\
		& V_t = \beta \hat{V}_{t-1} + (1-\beta) S^\tau _t\\
        & \hat{V}_t = (1-O_t) V_t + V_r O_t\\
        & O_t = \Theta(V_t)
	\end{aligned}
\end{equation}
\end{small}
where $S$ and $S^\tau$ denote the state and temporal state, respectively; $O_t$ is the output spike at the $t$-th time step; $*$ represents the matrix multiplication; and $W_E$ is the learnable matrix that can be jointly optimized. 

\subsubsection{Decoders}
The decoding method used in SRL is usually rate decoding. The accuracy of this decoding method is proportional to the length of the time step, so it is difficult to strike a good trade-off between performance and latency. To improve the performance of decoding at low latency, we use an operation similar to adaptive encoding. We multiply the spike trains with a learnable matrix $W_D$ to compress them $T$ times in the time dimension:
\begin{small}
\begin{equation}\label{eq::Decoder}
	\left \{ \begin{aligned}
        Q &= Q^\tau * W_{VD} \\
        A &= A^\tau * W_{AD} 
	\end{aligned}
    \right .
\end{equation}
\end{small}
where $Q^\tau$ and $A^\tau$ are output spike trains of the value network and action network, respectively; $W_{VD}$ and $W_{AD}$ are decode matrices of the value decoder and action decoder; and $Q/A$ are the value function and actions selection of RL. Specially, for RL algorithms with non-AC architectures such as DQN, we only decode the $Q$ values.

\subsection{Direct Training with Surrogate Gradients}
Compared with SNNs obtained using conversion methods, directly trained SNNs tend to have lower latency and higher inference energy efficiency. However, the non-differentiable property of spikes makes directly training SNNs difficult. The surrogate gradient method solves this problem by using alternative activation functions and is widely used to train deep SNNs \cite{wu2018spatio}. We adopt the arctangent function as the surrogate gradient function:
\begin{small}
\begin{align}
    \begin{split}
        \Theta'(x) &\triangleq h'(x) = \frac{\alpha}{2[1 + (\frac{\pi}{2}\alpha x)^2]}
    \end{split}
\end{align}
\end{small}
where $\alpha$ is a hyperparameter that controls the width of the surrogate function.

Suppose the loss function is $\mathcal{L}$, according to Eq.~\ref{eq::Decoder} and the chain rule, for the decoding matrix, we have:
\begin{small}
\begin{equation}
	\left\{
		\begin{aligned}
			\frac{\partial \mathcal{L}_Q}{\partial W_{VD}}= \frac{\partial \mathcal{L}_Q}{\partial Q}\frac{\partial Q}{\partial W_{VD}}=\frac{\partial \mathcal{L}_Q}{\partial Q}Q^\tau\\
			\frac{\partial \mathcal{L}_A}{\partial W_{AD}}=\frac{\partial \mathcal{L}_A}{\partial A}\frac{\partial A}{\partial W_{AD}}=\frac{\partial \mathcal{L}_A}{\partial A}A^\tau
		\end{aligned}
	\right.
\end{equation}
\end{small}
and for the final layer $L$:
\begin{small}
\begin{equation}
    \frac{\partial \mathcal{L}}{\partial W^L} = 
    \left \{ \begin{aligned}
        \frac{\partial \mathcal{L}_Q}{\partial Q} \frac{\partial Q}{\partial W^L}=\frac{\partial \mathcal{L}_Q}{\partial Q} \sum_{t=1}^T \frac{\partial Q^{\tau}_t}{\partial W^L} * W_{VD}\\
        \frac{\partial \mathcal{L}_A }{\partial A} \frac{\partial A}{\partial W^L}=\frac{\partial \mathcal{L}_A}{\partial A} \sum_{t=1}^T \frac{\partial A^{\tau}_t}{\partial W^L} * W_{AD}
	\end{aligned}
    \right .
\end{equation}
\end{small}
where $\frac{\partial \mathcal{L}_Q }{\partial Q}$ and $\frac{\partial \mathcal{L}_A }{\partial A}$ are defined according to different algorithms; $Q^{\tau}_t$ and $A^{\tau}_t$ are output spike trains, equal to $O^L_t$, so we only need to calculate $\frac{\partial O^L_t}{\partial W^L}$:
\begin{small}
\begin{equation}
    \begin{split}
        \frac{\partial O^L_t}{\partial W^L}&=\frac{\partial O^L_t}{\partial V^L_t} \frac{\partial V^L_t}{\partial W^L} = \frac{\partial \Theta(V^L_t)}{\partial V^L_t} (1-\beta) O^{l-1}_t\\
        &= h'(V^L_t) (1-\beta) O^{l-1}_t
    \end{split}
\end{equation}
\end{small}
and for the hidden layers $l \in \{1, \cdots, L-1\}$, according to Eq.~\ref{eq::LIF}:
\begin{small}
\begin{equation}
    \label{eq::Loss1}
    \frac{\partial \mathcal{L}}{\partial W^l} = \sum_{t=1}^T \frac{\partial \mathcal{L}}{\partial O^l_t} \frac{\partial O^l_t}{\partial V^l_t} \frac{\partial V^l_t}{\partial W^l}
\end{equation}
\end{small}
where $T$ is the total time steps. 

The first factor in Eq.~\ref{eq::Loss1} could be derived as:
\begin{small}
\begin{equation}
    \label{eq::pLpO}
    \begin{split}
        \frac{\partial \mathcal{L}}{\partial O^l_t} &= \frac{\partial \mathcal{L}}{\partial O^{l+1}_t} \frac{\partial O^{l+1}_t}{\partial V^{l+1}_t} \frac{\partial V^{l+1}_t}{\partial O^l_t}\\
        &=\frac{\partial \mathcal{L}}{\partial O^{l+1}_t} h'(V^{l+1}_t) (1-\beta) W^{l+1}
    \end{split}
\end{equation}
\end{small}
The second factor in Eq.~\ref{eq::Loss1} could be derived as:
\begin{small}
\begin{equation}
    \label{eq::pOpV}
    \begin{split}
        \frac{\partial O^l_t}{\partial V^l_t} = \frac{\partial \Theta (V^l_t)}{\partial V^l_t} = h'(V^l_t)
    \end{split}
\end{equation}
\end{small}
The third factor in Eq.~\ref{eq::Loss1} could be derived as:
\begin{small}
\begin{equation}
    \label{eq::V}
    \begin{split}
        \frac{\partial V^l_t}{\partial W^l} &= \beta \frac{\partial \hat{V}^l_{t-1}}{\partial W^l} + (1-\beta)O^{l-1}_t\\
    \end{split}
\end{equation}
\end{small}
where:
\begin{small}
\begin{equation}
    \label{eq::hatV}
    \begin{split}
        \frac{\partial \hat{V}^l_{t-1}}{\partial W^l} &= V_r \frac{\partial O^l_{t-1}}{\partial W^l} + (1-O^l_{t-1}) \frac{\partial V^l_{t-1}}{\partial W^l} - V^l_{t-1} \frac{\partial O^l_{t-1}}{\partial W^l}\\
        &=(V_r-V^l_{t-1})h'(V^l_{t-1})\frac{\partial V^l_{t-1}}{\partial W^l} + (1-O^l_{t-1}) \frac{\partial V^l_{t-1}}{\partial W^l}\\
        &=[(V_r-V^l_{t-1})h'(V^l_{t-1})+(1-O^l_{t-1})]\frac{\partial V^l_{t-1}}{\partial W^l}
    \end{split}
\end{equation}
\end{small}
Let:
\begin{small}
\begin{equation}
    \begin{split}
        \phi_t &\triangleq \beta [(V_r-V^l_{t-1})h'(V^l_{t-1})+(1-O^l_{t-1})]\frac{\partial V^l_{t-1}}{\partial W^l}\\
        \zeta_t &\triangleq (1-\beta)O^{l-1}_t
    \end{split}
\end{equation}
\end{small}
Then:
\begin{small}
\begin{equation}
    \label{eq::pVpW}
    \begin{split}
        \frac{\partial V^l_t}{\partial W^l} &= \phi_t \frac{\partial V^l_{t-1}}{\partial W^l} + \zeta_t \\
        &= \zeta_0 \prod_{i=1}^t \phi_i + \sum_{j=1}^{t-1} \zeta_j \prod_{k=j+1}^t \phi_k
    \end{split}
\end{equation}
\end{small}

Bring Eq.~\ref{eq::pLpO}, Eq.~\ref{eq::pOpV} and Eq.~\ref{eq::pVpW} into Eq.~\ref{eq::Loss1}, we can get the derivative of the loss functions in the hidden layers:
\begin{small}
    \begin{equation}
        \label{eq::loss}
        \begin{split}
            \frac{\partial \mathcal{L}}{\partial W^l} = \sum_{t=1}^T &\frac{\partial \mathcal{L}}{\partial O^{l+1}_t} h'(V^{l+1}_t) (1-\beta) W^{l+1} h'(V^l_t) \\
            &\cdot (\zeta_0 \prod_{i=1}^t \phi_i + \sum_{j=1}^{t-1} \zeta_j \prod_{k=j+1}^t \phi_k)
        \end{split}
    \end{equation}
\end{small}

\begin{table*}[t]
	\centering
    \small
    \resizebox{1.0\textwidth}{!}{
    \begin{threeparttable}
	\begin{tabular}{c|c|c c|c c|c c|c c}
        \bottomrule
        \textbf{Method} & \textbf{Vanilla DQN}\tnote{1} & \multicolumn{2}{c}{\textbf{Convert DQN}\tnote{2}} & \multicolumn{2}{|c}{\textbf{DSQN}\tnote{3}} & \multicolumn{2}{|c}{\textbf{DSQN}\tnote{4}} & \multicolumn{2}{|c}{\textbf{ACSF}} \cr
        \textbf{Networks} & \textbf{DNN} & \multicolumn{2}{c|}{\textbf{SNN(\boldmath $T=500$)}} & \multicolumn{2}{c|}{\textbf{SNN(\boldmath $T=64$)}} & \multicolumn{2}{c|}{\textbf{SNN(\boldmath $T=8$)}} & \multicolumn{2}{c}{\textbf{SNN(\boldmath $T=4$)}}\cr
        \textbf{Scores} & \textbf{Reward} & \textbf{Reward} & \textbf{\%DQN} & \textbf{Reward} & \textbf{\%DQN} & \textbf{Reward} & \textbf{\%DQN} & \textbf{Reward} & \textbf{\%DQN}\cr
		\hline
		Atlantis & 567506.6 & 177034.0 & 31.2\% & 487366.7 & 85.9\% & \textbf{2481620.0} & \textbf{437.3} \% & 1009850.0 & 177.9\% \cr 
        Beam Rider & 4152.4 &\textbf{9189.3} & \textbf{221.3\%} & 7226.9 & 174.0\% & 5188.9 & 125.0\% & 7472.2 & 179.9\% \cr
        Boxing & 94.8 & 75.5 & 79.6\% & \textbf{95.3} & \textbf{100.5}\% & 84.4 & 89.0\% & 65.0 & 68.6\% \cr
        Breakout & 204.2 & 286.7 & 140.4\% & \textbf{386.5} & \textbf{189.3\%} & 360.8 & 176.7\% & 336.2 & 164.6\% \cr
        Crazy Climber & 124413.3 & 106416.0 & 85.5\% & 123916.7 & 99.6\% & 93753.3 & 75.4\% & \textbf{120310.0} & \textbf{96.7\%} \cr
        Gopher & 5925.3 & 6691.2 & 112.9\% & \textbf{10107.3} & \textbf{170.6\%} & 4154.0 & 70.1\% & 6210.0 & 104.8\% \cr
        Jamesbond & 323.3 & 521.0 & 161.2\% & 1156.7 & 357.8\% & 463.3 & 143.3\% & \textbf{1275.0} & \textbf{394.4\%} \cr
        Kangaroo & 8693.3 & \textbf{9760.0} & \textbf{112.3\%} & 8880.0 & 102.1\% & 6140.0 & 70.6\% & 6770.0 & 77.9\% \cr
        Krull & 6214.0 & 4128.2 & 66.4\% & 9940.0 & 160.0\% & 6899.0 & 111.0\% & \textbf{11969.1} & \textbf{192.6\%} \cr
        Name This Game & 2378.0 & 8448.0 & 355.3\% & \textbf{10877.0} & \textbf{457.4\%} & 7082.7 & 297.8\% & 7668.0 & 322.5\% \cr
        Pong & 20.9 & 19.8 & 94.7\% & 20.3 & 97.1\% & 19.1 & 91.4\% & \textbf{21.0} & \textbf{100.5}\% \cr
        Road Runner & 47856.6 & 41588.0 & 86.9\% & 48983.3 & 102.4\% & 23206.7 & 48.5\% & \textbf{54710.0} & \textbf{114.3\%} \cr
        Space Invaders & 1333.8 & \textbf{2256.8} & \textbf{169.2\%} & 1832.2 & 137.4\% & 1132.3 & 84.9\% & 1658.0 & 124.3\% \cr
        Tennis & 19.0 & 13.3 & 70.0\% & -1.0 & -5.3\% & -1.0 & -5.3\% & \textbf{17.1} & \textbf{90.0\%} \cr
        Tutankham & 150.5 & 157.1 & 104.4\% & 194.7 & 129.4\% & 276.0 & 183.4\% & \textbf{281.9} & \textbf{187.3\%} \cr
        VideoPinball & 215016.4 & 74012.1 & 34.4\% & 275342.8 & 128.1\% & 441615.2 & 205.4\% & \textbf{445700.7} & \textbf{207.3\%} \cr
        \hline
        \multicolumn{2}{c}{Total} & \multicolumn{2}{c}{1925.8\%} & \multicolumn{2}{c}{2486.2\%} & \multicolumn{2}{c}{2204.5\%} & \multicolumn{2}{c}{\textbf{2603.6\%}} \cr
        \multicolumn{2}{c}{Win/Tie/Loss} & \multicolumn{2}{c}{7/1/8} & \multicolumn{2}{c}{9/5/2} & \multicolumn{2}{c}{8/0/8} & \multicolumn{2}{c}{\textbf{10/3/3}} \cr
        \bottomrule
	\end{tabular}
    \begin{tablenotes}
        \footnotesize
        \item[1] Reproduction results of DQN \cite{mnih2015human}.
        \item[2] Converting DQN to SNNs \cite{tan2021strategy}.
        \item[3] Directly-trained Deep Spiking Q-Network \cite{liu2021human}.
        \item[4] Deep Spiking Q-Network with non-spiking neural model \cite{chen2022deep}.
    \end{tablenotes}
    \end{threeparttable}
    }
    \caption{Performance of different DSQN algorithms on Atari games.}
	\label{Tab::DQN}
\end{table*}
\begin{figure}[t]
	\centerline{\includegraphics[width=1.0\linewidth]{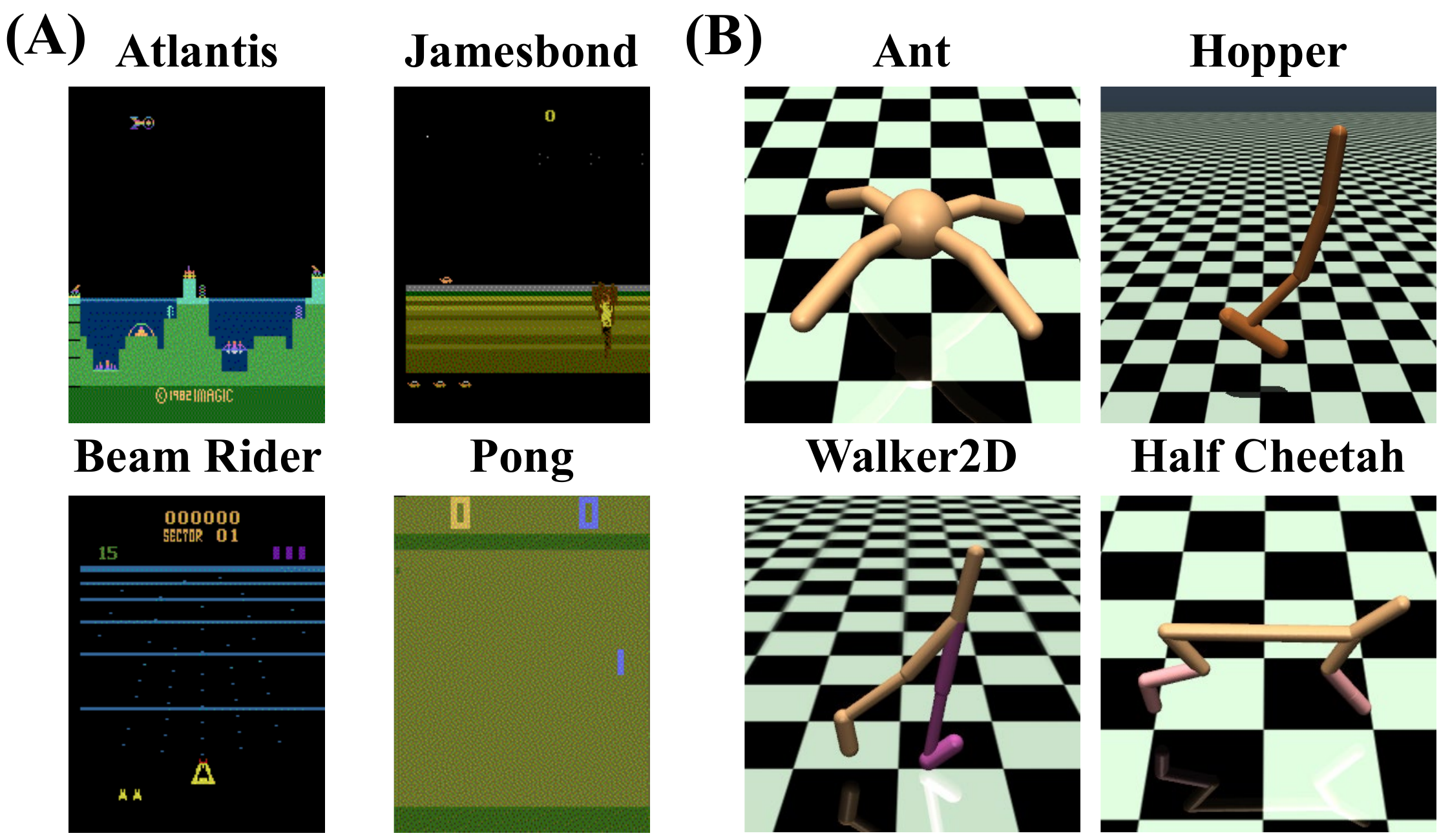}}
	\caption{Atari games and MuJoCo environments in the OpenAI gym. (A) Screenshots from various Atari games. The agent needs to be alive and earn more rewards. (B) MuJoCo robot control tasks, making robots of different shapes walk forward as fast as possible.}
	\label{Fig::Environments}
\end{figure}

\section{Experiments}
In this section, we apply the ACSF to both online and offline methods and evaluate them in different environments. We compared the ACSF with the latest SRL methods to prove that the ACSF can achieve a better trade-off between performance and latency. We also conduct ablation studies to demonstrate the superiority of adaptive coders.

\subsection{Experimental Settings}
\subsubsection{Environment}
To evaluate the proposed ACSF, we apply it to sixteen Atari 2600 games from the arcade learning environment (ALE) \cite{bellemare2013arcade} and four MuJoCo \cite{todorov2012mujoco} continuous robot control tasks from the OpenAI gym \cite{brockman2016openai}. We use the most stable version of the Atari game without frame skipping (NoFrameskip-v4) and the latest version of the MuJoCo environment (-v3). The visualization of environments is shown in Figure~\ref{Fig::Environments}. The training steps for Atari games and MuJoCo environments are 50 million (50M) and 1 million (1M), respectively, and the evaluation interval is 50 thousand (50K) and 5 thousand (5K).

\subsubsection{Training and Evaluating}
Baseline methods DQN, DDPG, BCQ, BC, and the proposed ACSF were all trained using the same structure as in \cite{fujimoto2019off}. Our results are reported over 5 random seeds \cite{DBLP:conf/icml/DuanCHSA16} of the behavioral policy, the OpenAI gym simulator, and network initialization. We save the model with the highest average reward for testing. 

We evaluate the ACSF on Atari games by playing 10 rounds of each game with an $\epsilon$-policy ($\epsilon=0.05$). Each episode is evaluated for up to 18K frames per round. To test the robustness of the agent in different situations, the agent starts with a random number (at most 30 times) of no-op actions. For MuJoCo environments, we played 10 rounds of each environment without exploration, and each test episode lasted for a maximum of 1K frames.
\begin{figure}[t]
	\centerline{\includegraphics[width=1.0\linewidth]{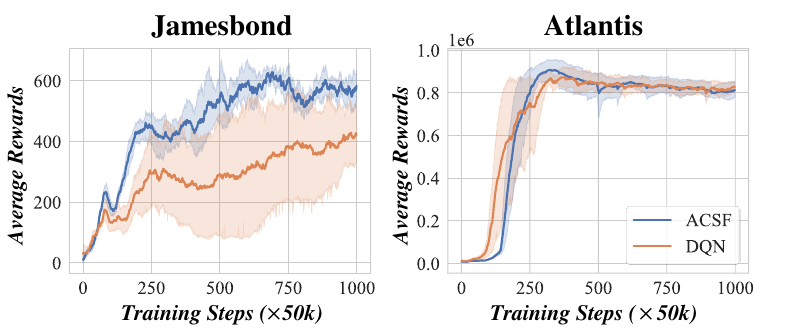}}
	\caption{Learning curves for DQN and ACSF. During the training process, the performance of ACSF meets or exceeds that of the DQN algorithm. The learning curves have been smoothed for aesthetics.}
	\label{Fig::DQN}
\end{figure}

\subsection{Results on Atari Games}
To test the performance of ACSF on game controls (a discrete action-space environment), we apply it to the DQN algorithm and test it on 16 different Atari games. We also compare the ACSF with some other DSQN methods \cite{tan2021strategy,liu2021human,chen2022deep} to show it can achieve an excellent trade-off between latency and performance.

The experimental results and comparisons are shown in Table~\ref{Tab::DQN} and Figure~\ref{Fig::DQN}. The full learning curves are included in the supplementary material. Experimental results show that ACSF outperforms the original DNQ algorithm. Meanwhile, compared to other SNN-based methods, ACSF maintains similar or better performance and reduces latency by more than 50\%. This also proves that adaptive coding performs better on RL tasks than fixed coding.

\subsection{Results on MuJoCo Robot Control Tasks}
\begin{figure*}[t]
	\centerline{\includegraphics[width=1.0\linewidth]{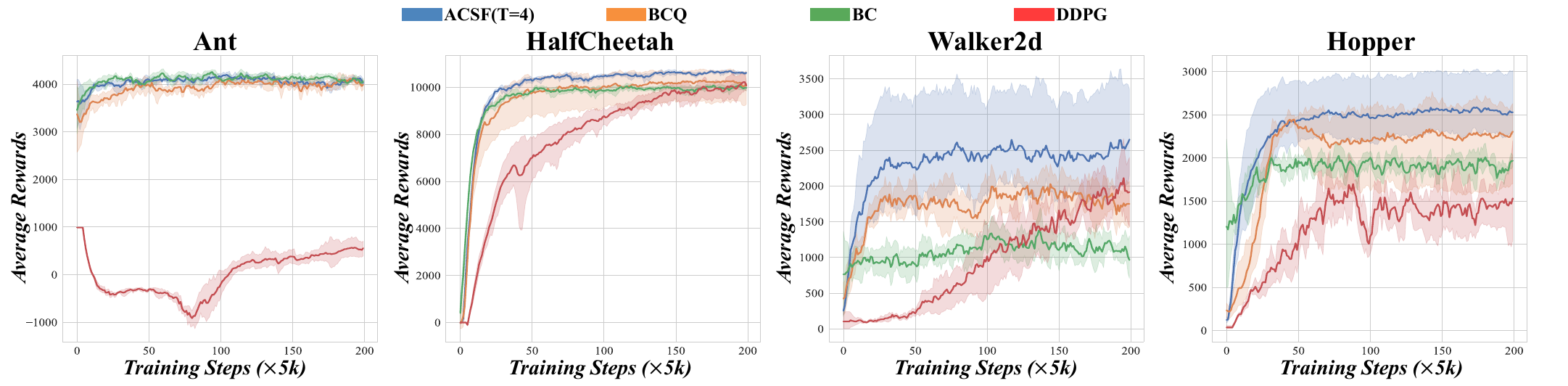}}
	\caption{Learning curves for different algorithms in the MuJoCo environment.}
	\label{Fig::BCQ}
\end{figure*}

The ACSF is a complete SRL framework that can be applied to both online and offline algorithms. Therefore, we apply ACSF to offline algorithms and conduct experiments in a discrete action-space environment to prove the versatility of ACSF. We compare the performance of ACSF with three baseline algorithms (DDPG, BC, BCQ). The results are shown in Table~\ref{Tab::BCQ} and Figure~\ref{Fig::BCQ}. From the results, ACSF is also suitable for offline algorithms, and the adaptive coders improve the performance significantly. Even at ultra-low latency ($T=4$), ACSF outperformed all the baseline algorithms. 

The performance of ACSF will also vary under different time step settings because the latency ($T$) affects the dimension of the learnable matrix in adaptive coders, and thus affects the fitting ability of coders. According to experimental results, ACSF can achieve its best performance when $T=8$ or $T=10$. We can select the value of $T$ flexibly according to the usage scenario.

\begin{table*}[t]
	\centering
    \small
	\begin{tabular}{c c c c c c c c}
        \bottomrule
        \textbf{Method} & \textbf{DDPG} & \textbf{BC} & \textbf{BCQ} & \textbf{ACSF} & \textbf{ACSF} & \textbf{ACSF} & \textbf{ACSF}\cr
        \textbf{Time Step} & \textbf{-} & \textbf{-} & \textbf{-} & \boldmath $T=4$ & \boldmath $T=6$ & \boldmath $T=8$ & \boldmath $T=10$ \cr
		\hline
		Ant & 542 $\pm$ 192 & 4052 $\pm$ 28 & 3959 $\pm$ 546 & 4100 $\pm$ 323 & 4526 $\pm$ 152 & \textbf{4598 $\pm$ 106} & 4567 $\pm$ 140 \cr
        HalfCheetah & 10066 $\pm$ 469 & 9978 $\pm$ 128 & 8430 $\pm$ 152 & 10610 $\pm$ 63 & 10960 $\pm$ 229 & \textbf{10983 $\pm$ 215} & 10913 $\pm$ 292 \cr
        Walker2d & 1917 $\pm$ 380 & 1103 $\pm$ 161 &  1770 $\pm$ 277 & 2551 $\pm$ 932 & 3545 $\pm$ 1680 & 4103 $\pm$ 1728 & \textbf{4589 $\pm$ 421} \cr
        Hopper & 1450 $\pm$ 320 & 1888 $\pm$ 105 & 2261 $\pm$ 603 & 2527 $\pm$ 455 & 2805 $\pm$ 919 & 3024 $\pm$ 663 & \textbf{3211 $\pm$ 639}\cr
        \bottomrule
	\end{tabular}
    \caption{Average reward for MuJoCo environments.}
	\label{Tab::BCQ}
\end{table*}

\subsection{Ablation Study}
To verify that adaptive coders can improve performance effectively. We conducted ablation experiments to compare the performance of algorithms with different coding methods. We compared five settings:
\begin{itemize}
    \item No coders+DNN: The original BCQ algorithm.
    \item Rate coders+SNN: Compute the firing rate of spikes as the value function.
    \item Accumulate coders+SNN: Modify the dynamics of the output neurons to make them stop firing spikes. Record the voltage at the last time step as the value function.
    \item Adaptive coders+DNN: On the basis of the original DNN, adaptive coders are added to eliminate the bias caused by the structural differences introduced by the adaptive coders. 
    \item Adaptive coders+SNN: Combines adaptive coders and directly-trained SNNs (the proposed ACSF).
\end{itemize}

\begin{table}[t]
	\centering
    \resizebox{0.5\textwidth}{!}{
    \begin{threeparttable}
	\begin{tabular}{c c c c c}
        \bottomrule
        \textbf{Settings} & \textbf{Ant} & \textbf{HalfCheetah} & \textbf{Walker2d} & \textbf{Hopper}\cr
		\hline
		None+DNN & 3959 $\pm$ 546 & 8430 $\pm$ 152 & 1770 $\pm$ 277 & 2261 $\pm$ 603\cr
        Rate+SNN\tnote{\dag} & -498 $\pm$ 155 & -95 $\pm$ 103 & 251 $\pm$ 40 & 99 $\pm$ 6\cr
        Accum.+SNN\tnote{\dag} & 3993 $\pm$ 490 & 9854 $\pm$ 196 & 1942 $\pm$ 416 & 2459 $\pm$ 259\cr
        \hline
        Adapt.+DNN\tnote{\dag} & 4076 $\pm$ 313 & 9960 $\pm$ 124 & 2080 $\pm$ 567 & 2268 $\pm$ 247\cr
        \textbf{Adapt.+SNN}\tnote{\dag} & \textbf{4100 $\pm$ 323} &\textbf{10610 $\pm$ 63} & \textbf{2551 $\pm$ 932} & \textbf{2527 $\pm$ 455}\cr
        \bottomrule
	\end{tabular}
    \begin{tablenotes}
        \footnotesize
        \item[\dag] The time step is set to $T=4$.
    \end{tablenotes}
    \end{threeparttable}
    }
    \caption{The performance of different coders.}
	\label{Tab::Ablation}
\end{table}

Table~\ref{Tab::Ablation} displays the results. Rate encoding does not work well due to very short time steps. The accumulate coders lose their spike characteristics, so the performance is better and similar to that of pure DNN. The combination of adaptive coders and DNN is slightly superior to pure DNN, which also proves that adaptive coders can integrate state information and calculate the value function accurately. The combination of adaptive coders and SNN has optimal performance in different environments, demonstrating the effectiveness of ACSF in low-latency settings.

\subsection{Power Estimation and Comparison}
To calculate the power consumption of SNNs on hardware, we count the total synaptic operations (SynOps) \cite{wu2021tandem}. It is defined as follows:
\begin{small}
\begin{equation} 
    \begin{split}
        SynOps=\sum_{t=1}^T \sum_{l=1}^{L-1} \sum_{j=1}^{N^l} f_{out,j}^l s_j^l[t]
    \end{split}
\end{equation}
\end{small}
where $f_{out}$ denotes the number of outgoing connections to the subsequent layer, $T$ denotes the total time steps, $L$ is the total number of layers, $N^l$ means the number of neurons in layer $l$, and $s$ denotes the spikes.

For DNNs:
\begin{small}
\begin{equation}
    \begin{split}
        SynOps=\sum_{l=1}^{L} f_{in}^l N^l
    \end{split}
\end{equation}
\end{small}
where $f_{in}^l$ represents the number of incoming connections to each neuron in layer $l$.
\begin{table}[t]
    \small
	\centering
	\begin{tabular}{c c c c c}
        \bottomrule
        \textbf{Algorithm} & \multicolumn{2}{c}{\textbf{DQN}} & \multicolumn{2}{c}{\textbf{BCQ}} \cr
        \textbf{Framework} & \textbf{DNN} & \textbf{ACSF} & \textbf{DNN} & \textbf{ACSF} \cr
        \hline
        SynOps &1.6M&\textbf{288.8K}&413.7K&\textbf{277.9K} \cr
        Energy Saved & - & \textbf{81.95\%} & - & \textbf{32.83\%} \cr
        \bottomrule
	\end{tabular}
    \caption{Total synaptic operations.}
	\label{Tab::PowerEstimation}
\end{table}

The total SynOps are shown in Table~\ref{Tab::PowerEstimation}. The ACSF improves energy efficiency (up to 5X) on both algorithms, providing a more efficient application method for DRL.

\section{Conclusion}
In this paper, we propose a low latency adaptive coding spike framework (ACSF) for SRL. The ACSF solves the high latency problem in SRL by using adaptive coding. At the same time, ACSF has expanded the application range of SRL by designing structures for online and offline algorithms. In terms of versatility, ACSF is the first model to achieve compatibility with both online and offline algorithms and achieve optimal performance in both discrete and continuous environments. In addition, ACSF uses a directly trained SNN, which does not require DNNs, so it can be easily migrated to neuromorphic hardware. Extensive experiments show that ACSF can achieve the same or higher performance with ultra-low latency compared to other SRL methods. Meanwhile, compared with traditional DRL methods, ACSF has a significant improvement on inference energy efficiency (up to 5X) while maintaining better performance. 

\section*{Acknowledgements}
This work was supported by National Key Research and Development Program of China under Grant No. 2020AAA0105900, National Natural Science Foundation of China under Grant No. 62236007 and U2030204 and the Key Research Project of Zhejiang Lab under Grant No. 2021KC0AC01.
\bibliographystyle{named}
\bibliography{ijcai23}

\end{document}